\documentclass[letterpaper, 10 pt, conference]{ieeeconf}  % Comment this line out if you need a4paper

\IEEEoverridecommandlockouts                              % This command is only needed if 
\usepackage{amsmath} % assumes amsmath package installed
\usepackage{amssymb}  % assumes amsmath package installed
\usepackage{overpic}
\usepackage{xcolor}

\usepackage{multirow}
\usepackage{booktabs}
\usepackage{subcaption}
\usepackage{tikz}
\usepackage{dsfont}

\usepackage[pdfauthor={Fabian K\"uppers, Jan Kronenberger, Jonas Schneider, Anselm Haselhoff},
			pdftitle={Bayesian Confidence Calibration for Epistemic Uncertainty Modelling}]{hyperref}

\title{\LARGE \bf
Bayesian Confidence Calibration for Epistemic Uncertainty Modelling
}

\definecolor{confidence}{rgb}{0.12, 0.46, 0.71}
\definecolor{ece}{rgb}{0.84, 0.15, 0.16}

\newcommand{\pdf}{f}

\newcommand{\prob}{Pr}
\newcommand{\probscore}{p}
\newcommand{\probmodel}{\pi}
\newcommand{\priormodel}{\pi}
\newcommand{\posteriormodel}{\pi}
\newcommand{\estimatedmodel}{\hat{\pi}}

\newcommand{\credibleinterval}{C}

\newcommand{\mean}{\mu}

\newcommand{\forwardpasses}{T}
\newcommand{\quantile}{\tau}

\newcommand{\inputvariate}{X}
\newcommand{\classvariate}{Y}
\newcommand{\predclassvariate}{\hat{Y}}
\newcommand{\matchedvariate}{M}
\newcommand{\predvariate}{\hat{P}}
\newcommand{\bboxvariate}{R}
\newcommand{\predbboxvariate}{\hat{R}}
\newcommand{\collectvariate}{S}

\newcommand{\inputset}{\mathcal{\inputvariate}}
\newcommand{\classset}{\mathcal{Y}}

\newcommand{\bboxset}{\mathcal{\bboxvariate}}

\newcommand{\class}{y}

\newcommand{\matched}{{m}}
\newcommand{\pred}{\hat{\probscore}}
\newcommand{\bbox}{r}

\newcommand{\collect}{s}

\newcommand{\numcollectedvariates}{K}

\newcommand{\numvariates}{J}

\newcommand{\numclasses}{K}

\newcommand{\numbins}{M}

\newcommand{\numsamples}{N}
\newcommand{\indexsamples}{i}

\newcommand{\parameter}{\theta}
\newcommand{\estimatedparameter}{\hat{\parameter}}
\newcommand{\allparameters}{\Theta}

\newcommand{\model}{h}
\newcommand{\calmodel}{g}

\newcommand{\calibrated}{\hat{q}}
\newcommand{\calibratedvariate}{\hat{Q}}

\newcommand{\ind}{\mathds{1}}

\newcommand{\centerx}{c_x}
\newcommand{\centery}{c_y}
\newcommand{\width}{w}
\newcommand{\height}{h}

\graphicspath{{img/}}
\newcommand*\circled[1]{\tikz[baseline=(char.base)]{\node[shape=circle,draw,inner sep=1pt] (char) {#1};}}

\author{Fabian K\"uppers$^{1}$, Jan Kronenberger$^{1}$, Jonas Schneider$^{2}$ and Anselm Haselhoff$^{1}$% <-this % stops a space
\thanks{$^{1}$Ruhr West University of Applied Sciences, Bottrop, Germany
	{\tt\small \{fabian.kueppers, jan.kronenberger, anselm.haselhoff\}@hs-ruhrwest.de}}%
\thanks{$^{2}$Elektronische Fahrwerksysteme GmbH, Gaimersheim, Germany
	{\tt\small jonas.schneider@efs-auto.com}}%
}

\usepackage[pscoord]{eso-pic}% The zero point of the coordinate systemis the lower left corner of the page (the default).

\newcommand{\placetextbox}[3]{% \placetextbox{<horizontal pos>}{<vertical pos>}{<stuff>}
	\setbox0=\hbox{#3}% Put <stuff> in a box
	\AddToShipoutPictureBG*{% Add <stuff> to current page background
		\put(\LenToUnit{#1\paperwidth},\LenToUnit{#2\paperheight}){\vtop{{\null}\makebox[0pt][c]{#3}}}%
	}%
}%

\begin{document}

\placetextbox{0.5}{0.05}{%
	%\\
	\parbox{\textwidth}{
		\scriptsize
		Paper has been accepted to Intelligent Vehicles Symposium (IV) 2021. \\
		\copyright 2021 IEEE. Personal use of this material is permitted. Permission from IEEE must be obtained for all other uses, in any current or future media, including reprinting/republishing this material for advertising or promotional purposes, creating new collective works, for resale or redistribution to servers or lists, or reuse of any copyrighted component of this work in other works.
	}
}%

\newpage
\maketitle
\thispagestyle{empty}
\pagestyle{empty}

%%%%%%%%%%%%%%%%%%%%%%%%%%%%%%%%%%%%%%%%%%%%%%%%%%%%%%%%%%%%%%%%%%%%%%%%%%%%%%%%
\begin{abstract}
	Modern neural networks have found to be miscalibrated in terms of confidence calibration, i.e., their predicted confidence scores do not reflect the observed accuracy or precision. Recent work has introduced methods for post-hoc confidence calibration for classification as well as for object detection to address this issue.
	Especially in safety critical applications, it is crucial to obtain a reliable self-assessment of a model.
	But what if the calibration method itself is uncertain, e.g., due to an insufficient knowledge base?
	
	We introduce Bayesian confidence calibration - a framework to obtain calibrated confidence estimates in conjunction with an uncertainty of the calibration method.
	Commonly, Bayesian neural networks (BNN) are used to indicate a network's uncertainty about a certain prediction. BNNs are interpreted as neural networks that use distributions instead of weights for inference. We transfer this idea of using distributions to confidence calibration. For this purpose, we use stochastic variational inference to build a calibration mapping that outputs a probability distribution rather than a single calibrated estimate.
	Using this approach, we achieve state-of-the-art calibration performance for object detection calibration. Finally, we show that this additional type of uncertainty can be used as a sufficient criterion for covariate shift detection. All code is open source and available at \textit{https://github.com/EFS-OpenSource/calibration-framework}.
\end{abstract}

%%%%%%%%%%%%%%%%%%%%%%%%%%%%%%%%%%%%%%%%%%%%%%%%%%%%%%%%%%%%%%%%%%%%%%%%%%%%%%%%

\section{Introduction}
Modern neural networks output a score attached to each decision. Ideally, this score can be interpreted as the network's confidence in its prediction, indicating the probability of correctness \cite{Guo2018}. However, it is a well known issue that these confidence scores neither reflect the actual observed accuracy in classification \cite{Niculescu2005, Naeini2015, Guo2018} nor the observed precision in object detection \cite{Kueppers2020}. 
If a deviation between predicted model scores and observed frequency is detected, a model is called \textit{miscalibrated}. Especially in safety-critical applications like autonomous driving or medical diagnosis, well-calibrated confidence estimates are crucial.
Several research has focused on improving confidence calibration either for classification \cite{Naeini2015, Kull2017, Guo2018, Kueppers2020} or more recently for object detection \cite{Neumann2018, Kueppers2020}. Those calibration methods map an uncalibrated confidence estimate to a calibrated one. 
In the same way as neural networks, a calibration mapping must also be trained using a separate data set.
But what if such a mapping needs to calibrate predictions which it is unsure about, e.g. if a sample is out of training distribution? For example, a driver assistance system for pedestrian recognition may detect a person in an image area that has not been covered by the training set of the calibration model. With position-dependent calibration \cite{Kueppers2020}, this can lead to a misleading confidence estimate that may affect the system's behavior.
Therefore, it is desirable for a calibration method to indicate whether a calibrated estimate is reliable.

For this reason, we introduce Bayesian confidence calibration.
Similar to Bayesian neural networks, our approach utilizes the idea of placing distributions over the weights of a calibration mapping. Using this approach, it is possible to quantify the intrinsic or epistemic uncertainty of the calibration mapping itself. We treat the model parameters in a Bayesian way using stochastic variational inference (SVI) where we replace each weight by a normal distribution.
Thus, we do not obtain a single calibrated estimate for a single prediction but rather a sample distribution indicating the epistemic uncertainty about the current prediction. This additional type of uncertainty might be used in conjunction with the calibrated confidence estimate to reliably reflect the observed frequency or to even reject a sample if necessary.
Our concept as well as a qualitative example are shown in Fig. \ref{img:introduction:qualitative} and Fig. \ref{img:introduction:concept}, respectively.
\begin{figure}[t]
	\centering
	\begin{overpic}[width=1.0\linewidth]{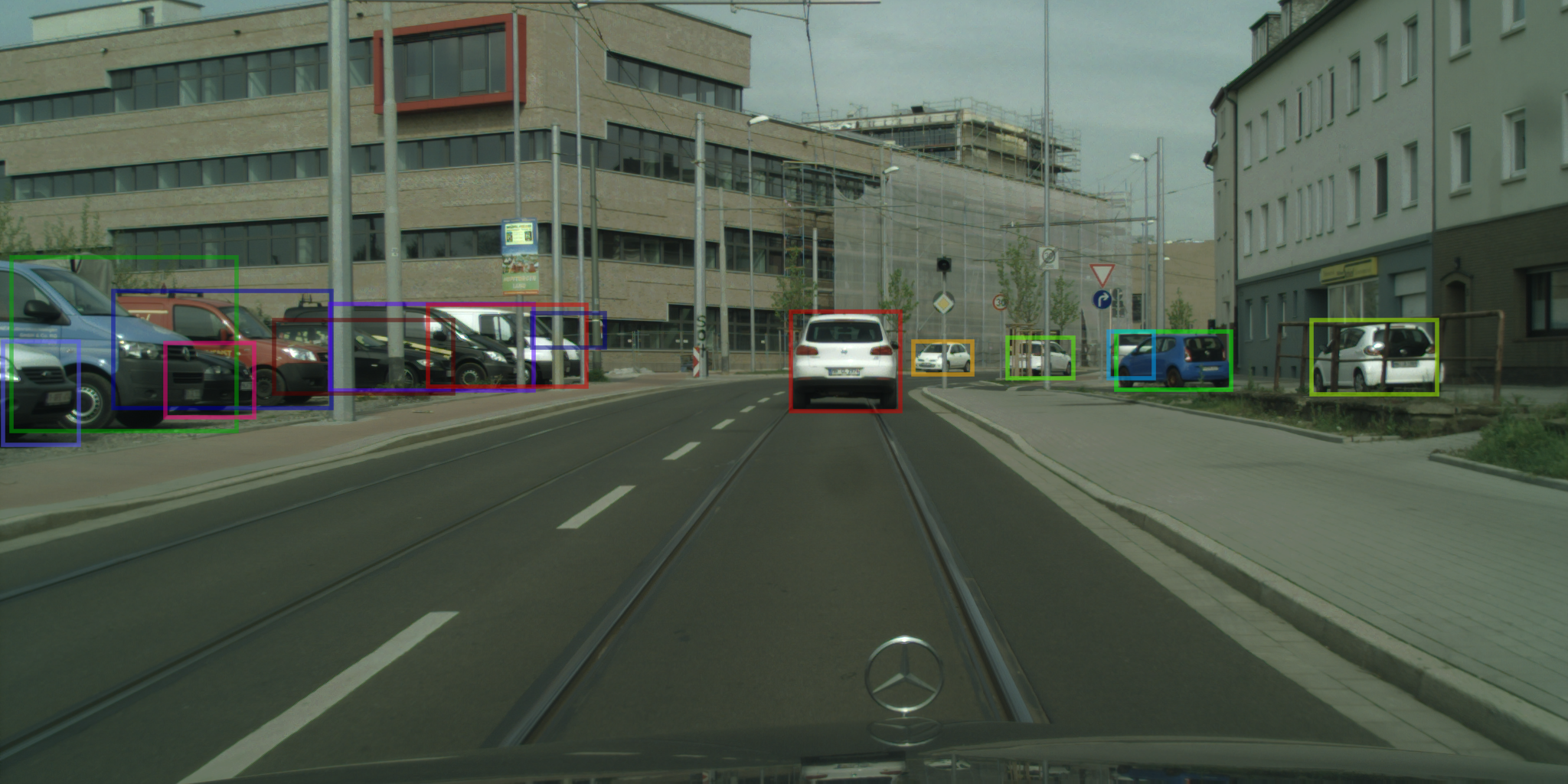}
		\put(52,31.5){\textbf{\scriptsize\color{red}\circled{1}}}
		\put(85,31){\textbf{\scriptsize\color{cyan}\circled{2}}}
		\put(64.5,30){\textbf{\scriptsize\color{orange}\circled{3}}}
		\put(0, 10){
			\colorbox{white}{
				\begin{minipage}[l]{8.8em}
					%{\small{\color{red}\circled{1}} $100\% \rightarrow 98\%${\raisebox{0.5ex}{\tiny$\substack{+1.2\% \\ -1.3\%}$}}}\\
					%{\small{\color{cyan}\circled{2}} $100\% \rightarrow 96\%${\raisebox{0.5ex}{\tiny$\substack{+3.8\% \\ -2.7\%}$}}}\\
					%{\small{\color{orange}\circled{3}} \hfill $99\% \rightarrow 85\%${\raisebox{0.5ex}{\tiny$\substack{+15.0\% \\ -14.8\%}$}}}
					\setlength{\tabcolsep}{1pt}
					\begin{tabular}{crrl}
						{\small \color{red}\circled{1}} & \small $100\% \rightarrow$ & \small $98\%$ & \small {\raisebox{0.5ex}{\tiny$\substack{+1.2\% \\ -1.3\%}$}}\\
						{\small \color{cyan}\circled{2}} & \small $100\% \rightarrow$ & \small $96\%$ & \small {\raisebox{0.5ex}{\tiny$\substack{+3.8\% \\ -2.7\%}$}}\\
						{\small \color{orange}\circled{3}} & \small $99\% \rightarrow$ & \small $85\%$ & \small {\raisebox{0.5ex}{\tiny$\substack{+15.0\% \\ -14.8\%}$}}
					\end{tabular}
				\end{minipage}
			}
		}
	\end{overpic}
	\caption{Using stochastic variational inference (SVI), we are able to obtain not only a single calibrated estimate but also an additional prediction interval quantifying the epistemic uncertainty within the calibration mapping. We use the position-dependent calibration framework of \cite{Kueppers2020} but place distributions over the calibration parameters to infer a sample distribution for a single prediction.}
	\label{img:introduction:qualitative}
\end{figure}
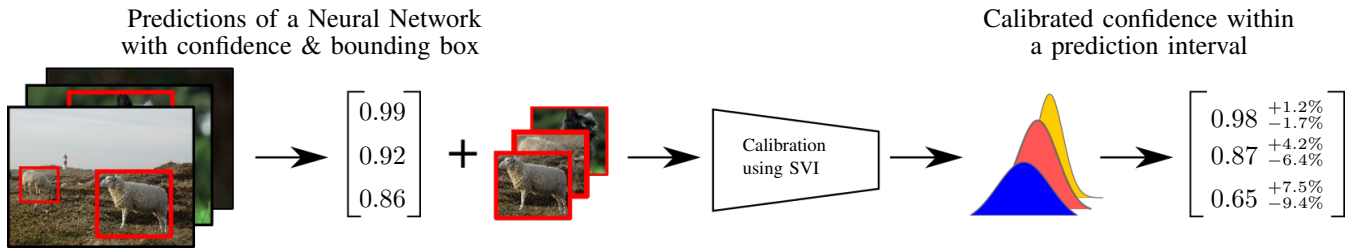
\begin{figure*}[t]
	\centering
	\begin{overpic}[width=1.0\linewidth]{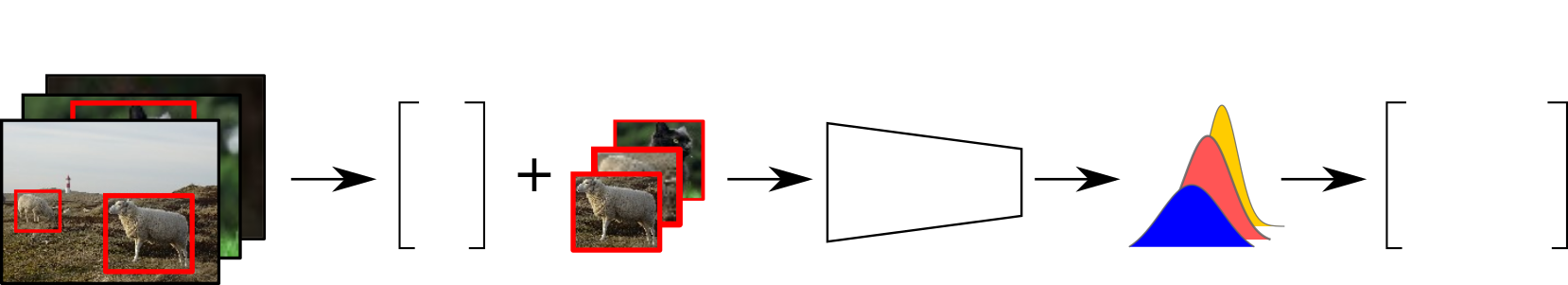}
		\put(9,16.5){Predictions of a Neural Network}
		\put(8.5, 14.5){with confidence \& bounding box}
		\put(73,16.5){Calibrated confidence within }
		\put(76.5,14.5){a prediction interval}
		
		%\put(53.8, 10.7){\small{Calibration}}
		\put(55, 7.2){\scriptsize{Calibration}}
		\put(55, 5.5){\scriptsize{using SVI}}
		%\put(53, 3.8){\scriptsize{using SVI}}
		
		\put(26.3,9.5){$0.99$}
		\put(26.3,6.25){$0.92$}
		\put(26.3,3){$0.86$}
		
		\put(90,9.){$0.98$ {\raisebox{0.5ex}{\small$\substack{+1.2\% \\ -1.7\%}$}}}
		\put(90,6.25){$0.87$ {\raisebox{0.5ex}{\small$\substack{+4.2\% \\ -6.4\%}$}}}
		\put(90,3){$0.65$ {\raisebox{0.5ex}{\small$\substack{+7.5\% \\ -9.4\%}$}}}
	\end{overpic}
	\caption{An object detection model outputs a confidence estimate attached to each bounding box with a certain position and shape. This information is used for position \& scale dependent confidence calibration \cite{Kueppers2020}. Instead of maximum likelihood estimation, we utilize stochastic variational inference to predict a sample distribution for each detection. On the one hand, this sample distribution reflects the observed frequency and on the other hand the epistemic uncertainty of the calibration model for a certain confidence, position and shape. Using highest density interval estimation, it is thus possible to denote a prediction interval for each calibrated estimate.}
	\label{img:introduction:concept}
\end{figure*}

This work is structured as follows: first, we give a review of the recent advances in confidence calibration and Bayesian uncertainty modelling. Second, we introduce our concept of using SVI within a calibration mapping and provide extensive studies using several pretrained network architectures. Finally, we conclude our contributions and show further research directions that arise from this work.\\

\textbf{Novelty and significance.} We introduce the concept of Bayesian confidence calibration. This is the first work discussing epistemic uncertainty within the scope of confidence calibration to the best of our knowledge. Furthermore, the evaluation of uncertainty obtained by a Bayesian neural network is commonly restricted to regression tasks \cite{Kuleshov2018, Song2018, Pearce2018,Song2019, Ding2020}. In contrast, we discuss how to evaluate epistemic uncertainty obtained by a Bayesian model within the scope of object detection. Therefore, this work has a considerable significance for safety-critical applications (e.g. autonomous driving, medical diagnosis, etc.) where a reliable uncertainty measure is crucial.
\section{Related Work}
\textbf{Confidence calibration.} Most calibration methods are applied to predictions of neural networks after inference as a post-processing step. We focus on scaling methods like logistic calibration aka Platt scaling \cite{Platt1999} and beta calibration \cite{Kull2017} where the logits of a network are scaled by learned parameters before applying a sigmoid/softmax. We further use histogram binning \cite{Zadrozny2001} as a baseline representative of the binning calibration methods. Recent work has shown that modern object detectors also tend to be miscalibrated \cite{Neumann2018, Feng2019, Kueppers2020}. The authors in \cite{Kueppers2020} showed that miscalibration in object detection also depends on the position and scale of the predicted objects. They provide a natural extension to logistic calibration and to beta calibration to also include the regression branch into a calibration mapping. Our investigations are based on this framework. Additionally, the authors in \cite{Kueppers2020} introduce the \textit{detection expected calibration error} (D-ECE) that is an extension of the well-known \textit{expected calibration error} (ECE) \cite{Naeini2015}. The D-ECE computes the miscalibration not only by using confidence, but also by including the position and scale of the detected objects.
%A drawback of using the ECE has been addressed by \cite{Kumar2019, Wenger2020, Patel2020, Ding2020} where they show that using a dedicated binning scheme underestimates the true calibration error in many cases. Warum verwenden wir ihn dann trotzdem? Keine Alternative, beste verfügbare Metrik, lower-bound ECE nicht implementiert.

Other work investigates how to directly train a calibrated output distribution. The authors in \cite{Neumann2018} use an additional network output that is learned in conjunction with the remaining logits and used to rescale the output probability (similar to temperature scaling \cite{Guo2018}). In contrast, \cite{Mukhoti2020} use a focal loss to obtain calibrated estimates for classification. The authors in \cite{Pereyra2017} add a confidence penalty term to obtain lower confidence estimates. 
A more targeted approach to address miscalibration within model training is proposed by \cite{Seo2019} where the authors use multiple stochastic forward passes with dropout enabled during training. The authors show that the uncertainty is highly correlated to miscalibration. Seo et al. use the sample's variance to weight a dedicated regularization term. In contrast to these approaches, we focus on post-hoc calibration methods for object detection.\\

\textbf{Bayesian neural networks.} The most common way to model epistemic uncertainty in the scope of object detection is to treat a model as a Bayesian neural network \cite{Graves2011, Blundell2015, Gal2016}. In the past, \cite{Kendall2017} proposed a framework to also model aleatoric uncertainty in conjunction with epistemic uncertainty. However, the number of parameters within a calibration method is highly limited, thus modelling aleatoric uncertainty is currently not feasible for a calibration mapping. Recently, \cite{Lakshminarayanan2017} showed that epistemic uncertainty estimates of BNNs are also miscalibrated, especially for the task of regression. 
A metric for uncertainty evaluation of a BNN has recently been introduced by \cite{Hall2020} named \textit{probability quality measure} (PDQ) \cite{Hall2020}. However, this metric is rather designed for spatial uncertainty evaluation for object detection or segmentation.
Extensive studies have focused on the calibration of uncertainty estimates obtained by Bayesian regression models \cite{Lakshminarayanan2017, Kuleshov2018, Song2018, Pearce2018, Song2019}. 
In most papers, perfect uncertainty calibration for regression is defined as the coverage probability of all prediction intervals containing the ground truth value (also called \textit{prediction interval coverage probability (PICP) \cite{Pearce2018}}). We further adapt this definition for uncertainty evaluation, too.\\
%Another metric called \textit{uncertainty calibration error} (UCE) proposed by \cite{Laves2019} measures the average entropy of the softmaxes within a bin and compares it to the classification error. This metric however is not applicable to evaluate uncertainty obtained by a BNN. 

\textbf{Improving confidence estimates.}
Orthogonal research directions are proposed by \cite{Jiang2018} (IoU-Net) and \cite{Rezatofighi2019} (GIoU), respectively. In contrast to common confidence calibration, IoU-Net focuses on fine-tuning the regression output to improve the IoU between predicted and ground-truth position. These authors introduce an additional location confidence (regression) in conjunction with the confidence (categorical). They even support the need for confidence calibration because the authors show that the ’regular’ confidence is not a direct measure for misalignment.
One of the most related approaches to our work has been presented by \cite{Allikivi2019} which introduces a non-parametric Bayesian isotonic calibration method. A prior distribution is used to sample many isotonic regression models. In contrast to our approach, the authors do not model epistemic uncertainty but rather use the likelihood of each sampled model in order to perform Bayesian model averaging.\\
%These works are fruitful Ergänzung zu unserer Arbeit. Die untersuchen das Modell direkt, während wir uns eher auf die Kalibrierungsmethoden selbst beschränken. Wenn bei denen dies im Kontext von BNNs gemacht wird, könnte das eine interessante Ergänzung werden.

\section{Confidence calibration using variational inference}
\label{section:methods}

\textbf{Definitions.}
For object detection tasks, we interpret the input images $\inputvariate \in \inputset$, all class labels $\classvariate \in \classset = \{1, ..., \numclasses\}$ and the object positions $\bboxvariate \in \bboxset^\numvariates = [0, 1]^\numvariates$ attached to each object (where $\numvariates$ denotes the size of the box encoding) as random variables that follow a joint ground-truth distribution $\probmodel(\inputvariate, \classvariate, \bboxvariate) = \probmodel(\classvariate, \bboxvariate | \inputvariate) \priormodel(\inputvariate)$. Neural networks $\model$ serve as a mapping of the input $\inputvariate$ to certain labels $\predclassvariate$ with confidence levels $\predvariate \in [0, 1]$ and boxes $\predbboxvariate \in \bboxset^\numvariates$ so that $\model: \inputvariate \rightarrow (\predclassvariate, \predvariate, \predbboxvariate)$.
According to \cite{Guo2018}, we would expect that the predicted scores $\predvariate$ represent a probability of correctness, i.e., they should match the observed accuracy for a certain confidence level.
The term of calibration in the scope of object detection differs in the sense that the predicted confidence $\predvariate$ should reflect the observed \textit{precision} for a certain confidence level at a certain position and shape \cite{Kueppers2020}. An object detection model is perfectly calibrated if
\begin{align}
&\prob(\matchedvariate=1 | \predvariate=\probscore, \predclassvariate=\class, \predbboxvariate=\bbox) = \probscore \\
\nonumber &\forall \probscore \in [0,1], \class \in \classset, \bbox \in \bboxset^\numvariates
\end{align}
holds \cite{Kueppers2020}, where $\matchedvariate=1$ denotes a correctly detected prediction that matches a ground-truth object at a certain IoU score. If we observe a deviation between predicted confidence and observed precision, the model $\model$ is miscalibrated.
For classification, calibration methods aim to map a confidence score of an uncalibrated classifier to a calibrated one $\calibratedvariate \in [0, 1]$ that matches the accuracy for a certain confidence level. Such a calibration method $\calmodel$ is a post-processing method that needs an own training stage to learn a mapping $\calmodel: \predvariate \rightarrow \calibratedvariate$ with calibration parameters $\estimatedparameter$ and can be seen as a probabilistic model $\estimatedmodel(\classvariate | \predvariate, \estimatedparameter)$. Recently, the authors in \cite{Kueppers2020} have shown that calibration for object detection also depends on the regression output. They provide an extension to existing calibration methods to build a calibration mapping $\estimatedmodel(\matchedvariate | \predvariate, \predclassvariate, \predbboxvariate, \estimatedparameter)$ that also includes the regression output by $\calmodel: (\predvariate, \predclassvariate, \predbboxvariate) \rightarrow \calibratedvariate$. We denote $\collectvariate = (\predvariate, \predbboxvariate)$ and stick to the calibration only for a single class "pedestrian" so that our calibration methods are independent of $\classvariate$ in the following.\\

\textbf{Bayesian confidence calibration.}
For all scaling methods, the calibration parameters $\estimatedparameter$ are commonly obtained by maximum likelihood (ML) estimation by minimizing the NLL loss.
Instead, we place an uninformative Gaussian prior $\priormodel(\parameter)$ with high variance over the parameters $\parameter$ and infer the posterior given by
\begin{align}
\probmodel(\parameter | \collectvariate, \matchedvariate) = \frac{\probmodel(\matchedvariate | \collectvariate, \parameter)\priormodel(\parameter)} {\int_\allparameters \probmodel(\matchedvariate | \collectvariate, \parameter)\priormodel(\parameter) d\parameter}
\end{align}
with $\probmodel(\matchedvariate | \collectvariate, \parameter)$ as the likelihood. This distribution captures the most probable calibration parameters given the network output $\collectvariate$ and the according ground truth information $\matchedvariate$. Given this posterior, we can map a new input $\collect^\ast$ with the posterior predictive distribution defined by
\begin{align}
\pdf(\matched^\ast | \collect^\ast, \collectvariate, \matchedvariate) = \int_\allparameters \probmodel(\class^\ast | \collect^\ast, \parameter) \posteriormodel(\parameter | \collectvariate, \matchedvariate) d\parameter
\end{align}
to obtain a distribution as the calibrated estimate.
Since the posterior cannot be determined analytically, we use stochastic variational inference (SVI) \cite{Jordan1999, Hoffman2013, Gal2016a} as an approximation where a variational distribution (usually a Gaussian) is used whose structure is easy to evaluate. The parameters of the variational distribution are optimized to match the true posterior using the evidence lower bound (ELBO) loss \cite{Jordan1999, Gal2016a}. Afterwards, we sample $\forwardpasses$ sets of weights and use them to obtain a sample distribution consisting of $\forwardpasses$ estimates for a new single input $\collect^\ast$. %Finally, we use the mean as the new calibrated confidence estimate $\calibrated = \mean_{\pdf}$ with standard deviation $\stddev_{\pdf}$ as an additional uncertainty quantification for the calibrated estimate.

In contrast to Bayesian neural networks (BNNs), we model epistemic uncertainty of the calibration mapping. The distribution $\pdf_\indexsamples$ obtained by calibration for a sample with index $\indexsamples$ does not reflect the model uncertainty of a single prediction, but rather for a certain confidence level (and for a certain position/scale for object detection). For example, epistemic uncertainty estimates obtained by a BNN are computed for each input image separately. A prediction with a confidence score of $70\%$ might have a different uncertainty than another prediction for the same confidence level. In contrast, our approach assigns an equal uncertainty to both samples.\\

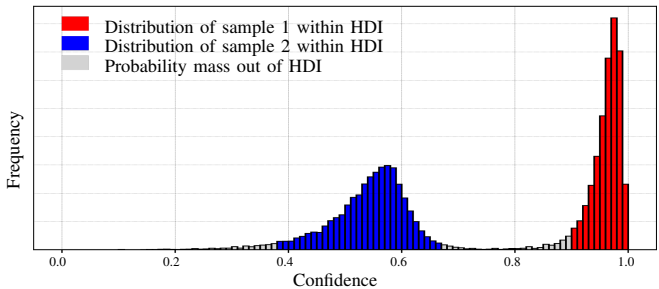
\begin{figure}[t]
	\centering
	\begin{overpic}[tics=5, width=1.0\linewidth]{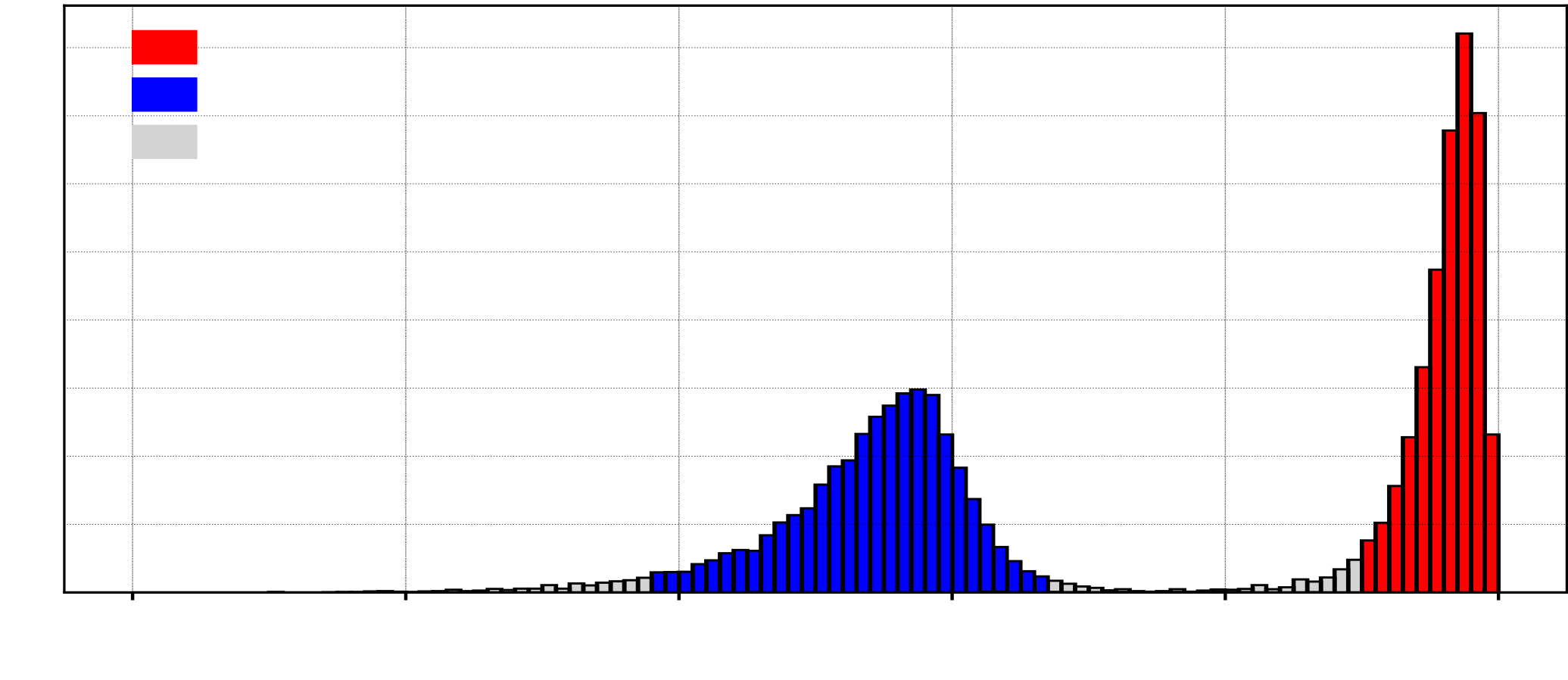}
		\put(44,0){\scriptsize{Confidence}}
		\put(0,15){\rotatebox{90}{\scriptsize{Frequency}}}
		
		\put(15,39){\scriptsize{Distribution of sample 1 within HDI}}
		\put(15,36){\scriptsize{Distribution of sample 2 within HDI}}
		\put(15,33){\scriptsize{Probability mass out of HDI}}
		
		\put(6.5,3){\tiny{0.0}}
		\put(24,3){\tiny{0.2}}
		\put(41.5,3){\tiny{0.4}}
		\put(59,3){\tiny{0.6}}
		\put(76.5,3){\tiny{0.8}}
		\put(94,3){\tiny{1.0}}
		
	\end{overpic}
	\caption{Exemplary confidence distributions of two predictions after logistic calibration using SVI. We propose to use the highest density interval (HDI) to get meaningful interval boundaries since the output distributions are mostly skewed.}
	\label{img:method:datadist}
\end{figure}

\textbf{Implications for uncertainty evaluation.}
Commonly, miscalibration in the scope of object detection is measured with the detection expected calibration error (D-ECE) \cite{Naeini2015, Kueppers2020}. We interpret the mean of the sample distribution as the new calibrated estimate $\calibrated=\mean_\pdf$ and use this to compute the D-ECE. The epistemic uncertainty can be expressed by a prediction interval around the mean estimate.
In frequentist statistics, it is quite common to choose quantile-based interval boundaries for a certain confidence level $\quantile$ assuming a normal distribution.
However, it is also possible that the sample distributions obtained by SVI do not follow a normal distribution (see Fig. \ref{img:method:datadist}).

Alternatively, in Bayesian terms, a prediction interval can also be described as a credible interval for the observed variable itself. Therefore, we use the \textit{highest density interval} (HDI) on the posterior predictive distribution to obtain the interval boundaries by 
\begin{align}
\credibleinterval_{\quantile, \indexsamples} = (\ell_\indexsamples, u_\indexsamples): \prob(\ell_\indexsamples \leq \text{prec}(i) \leq u_\indexsamples) = 1-\quantile ,
\end{align}
where prec$(\indexsamples)$ denotes the observed precision of sample $i$ for a certain $\collect_\indexsamples$. The advantage of using HDI is that it is possible to obtain the narrowest interval for a desired probability mass while being independent of the shape of the distribution. This is demonstrated in Fig. \ref{img:method:datadist} for two exemplary distributions obtained by SVI with their respective prediction intervals.

For uncertainty evaluation, we adapt the definition of quantile-calibrated regression \cite{Rueda2007, Gneiting2007, Kuleshov2018, Song2019, Fasiolo2020}. Given a calibration model $\calmodel$ that outputs a PDF $\pdf_\indexsamples$ for an input with index $\indexsamples$ out of $\numsamples$ samples, the uncertainty is well calibrated if the observed precision of all samples falls into a $100(1-\quantile)\%$ prediction interval (PI) approximately $100(1-\quantile)\%$ of the time. 

We can use the prediction intervals obtained by HDI to calculate the \textit{prediction interval coverage probability} (PICP) \cite{Pearce2018} which is defined by
\begin{align}
	\text{PICP} = \frac{1}{\numsamples}\sum^\numsamples_{\indexsamples=1} \ind(\text{prec}(i) \in \credibleinterval_{\quantile, \indexsamples}).
\end{align}
The definition of PICP is commonly used for calibrated regression where the true target value is known. However, for classification or object detection, the true precision is not directly accessible.
Therefore, we use a binning scheme over all available quantities $\collectvariate$ with $\numsamples$ samples to estimate the precision for each sample similar to the D-ECE calculation \cite{Kueppers2020}.
For perfect uncertainty calibration, it is required that $\text{PICP} \rightarrow (1-\quantile)$ as $\numsamples \rightarrow \infty$ \cite{Kuleshov2018, Song2019}. Using this definition, we can measure the difference between PICP and $(1-\quantile)$ to evaluate the uncertainty. As already mentioned by the authors of \cite{Kuleshov2018}, it is not sufficient for a probabilistic calibration model to output well-calibrated mean estimates. For example, a distribution with a wide prediction interval might be well calibrated in terms of the D-ECE or PICP but is also uninformative. Therefore, we also denote the \textit{mean prediction interval width} (MPIW) as a complementary measure where the prediction interval width for certain $\credibleinterval_{\quantile, \indexsamples}$ is averaged over all $\numsamples$ samples \cite{Pearce2018}.
Using D-ECE, PICP and MPIW, we are thus able to measure the quality of the calibration mapping itself as well as the quality of the epistemic uncertainty quantification.

\section{Experiments}
\begin{table*}[t!]
	\centering
	\caption{Calibration results of ML estimation with the respective differences to SVI estimation using multidimensional histogram binning (HB), conditional dependent logistic calibration (LC) and beta calibration (BC) \cite{Kueppers2020}. Each column of a single table denotes which subset of data has been used for calibration and for measuring the D-ECE. 
	Note that only scores of a single column can be compared to each other since each column uses a different binning to evaluate the miscalibration.
	}
	\label{result:table:calibration}
	\caption*{D-ECE [\%] on MS COCO validation set \cite{Lin2014} with IoU 0.50 (left) and 0.75 (right)}
	\begin{subtable}{0.48\textwidth}
		\centering
		\caption{Faster R-CNN X101-FPN \cite{Ren2015} with IoU 0.50}
		\resizebox{0.95\textwidth}{!}{%
			\begin{tabular}{lllll}
\toprule
{} &        $(\pred)$ &        $(\pred, \centerx, \centery)$ &          $(\pred, \width, \height)$ &          full \\
\midrule
baseline &         5.649 &         5.837 &         6.073 &         6.360 \\
HB       &         1.444 &         5.642 &         2.677 &         4.739 \\
LC       &  1.952 $+$0.085&  5.693 $-$0.011  &  2.320 $+$0.193 &  4.149 $+$0.125 \\
BC       &  1.584 $+$0.766&  5.691 $+$0.164 &  2.374 $+$0.220 &  4.245 $+$0.157  \\
\bottomrule
\end{tabular}

		}
	\end{subtable}
	\vspace{1em}
	\begin{subtable}{0.48\textwidth}
		\centering
		\caption{Faster R-CNN X101-FPN \cite{Ren2015} with IoU 0.75}
		\resizebox{0.95\textwidth}{!}{%
			\begin{tabular}{lllll}
\toprule
{} &        $(\pred)$ &        $(\pred, \centerx, \centery)$ &          $(\pred, \width, \height)$ &          full \\
\midrule
baseline &        13.295 &        12.660 &        12.744 &        12.395 \\
HB       &         1.469 &         6.606 &         3.664 &         5.672 \\
LC       &  2.972 $+$0.094 &  7.015 $+$0.399 &  3.682 $+$0.076 &  4.986 $+$0.059 \\
BC       &  2.345 $+$6.099 &  7.118 $+$0.186 &  3.358 $+$0.545 &  5.472 $-$0.082 \\
\bottomrule
\end{tabular}

		}
	\end{subtable}
	
	\begin{subtable}{0.48\textwidth}
		\centering
		\caption{RetinaNet R101-FPN \cite{Lin2017} with IoU 0.50}
		\resizebox{0.95\textwidth}{!}{%
			\begin{tabular}{lllll}
\toprule
{} &        $(\pred)$ &        $(\pred, \centerx, \centery)$ &          $(\pred, \width, \height)$ &          full \\
\midrule
baseline &        13.665 &        12.469 &        15.823 &        13.449 \\
HB       &         1.532 &         6.135 &         3.325 &         4.594 \\
LC       &  2.002 $+$0.034 &  6.149 $+$0.060 &  3.146 $+$0.600 &  4.120 $+$0.288 \\
BC       &  1.526 $-$0.047 &  6.332 $+$0.067 &  2.980 $+$0.193 &  4.713 $-$0.422 \\
\bottomrule
\end{tabular}

		}
	\end{subtable}
	\vspace{1.5em}
	\begin{subtable}{0.48\textwidth}
		\centering
		\caption{RetinaNet R101-FPN \cite{Lin2017} with IoU 0.75}
		\resizebox{0.95\textwidth}{!}{%
			\begin{tabular}{lllll}
\toprule
{} &        $(\pred)$ &        $(\pred, \centerx, \centery)$ &          $(\pred, \width, \height)$ &          full \\
\midrule
baseline &         2.988 &         7.935 &         8.270 &         7.251 \\
HB       &         1.689 &         7.152 &         4.061 &         5.412 \\
LC       &  1.970 $-$0.045 &  7.497 $+$0.030 &  3.963 $+$0.467 &  5.049 $-$0.032 \\
BC       &  1.743 $+$0.095 &  7.504 $+$0.133 &  4.002 $+$0.285 &  5.942 $-$0.598 \\
\bottomrule
\end{tabular}

		}
	\end{subtable}
	
	\caption*{D-ECE [\%] on Cityscapes validation set \cite{Cordts2016} with IoU 0.50 (left) and 0.75 (right)}
	\begin{subtable}{0.48\textwidth}
		\centering
		\caption{Mask-RCNN R50-FPN \cite{He2017} with IoU 0.50}
		\resizebox{0.95\textwidth}{!}{%
			\begin{tabular}{lllll}
\toprule
{} &        $(\pred)$ &        $(\pred, \centerx, \centery)$ &          $(\pred, \width, \height)$ &          full \\
\midrule
baseline &        10.502 &         8.663 &         9.842 &         9.974 \\
HB       &         3.022 &         4.501 &         3.158 &         5.366 \\
LC       &  3.418 $+$0.152 &  4.317 $+$0.556 &  2.366 $+$1.360 &  4.935 $+$0.655 \\
BC       &  3.572 $+$1.164 &  4.670 $-$0.073 &  2.694 $+$0.517 &  5.349 $+$0.219 \\
\bottomrule
\end{tabular}

		}
	\end{subtable}
	\begin{subtable}{0.48\textwidth}
		\centering
		\caption{Mask-RCNN R50-FPN \cite{He2017} with IoU 0.75}
		\resizebox{0.95\textwidth}{!}{%
			\begin{tabular}{lllll}
\toprule
{} &        $(\pred)$ &        $(\pred, \centerx, \centery)$ &          $(\pred, \width, \height)$ &          full \\
\midrule
baseline &        30.102 &        24.046 &        29.190 &        27.939 \\
HB       &         2.202 &         5.791 &         5.914 &         6.937 \\
LC       &  4.399 $+$0.232 &  5.887 $+$0.534 &  3.699 $+$0.928 &  5.945 $+$0.165 \\
BC       &  4.299 $+$4.962 &  6.021 $+$0.118 &  4.503 $+$0.198 &  6.259 $+$0.120 \\
\bottomrule
\end{tabular}

		}
	\end{subtable}
\end{table*}
\textbf{Experimental setup.}
We utilize the scaling methods \textit{logistic calibration} aka Platt scaling (LC) \cite{Platt1999} and \textit{beta calibration} (BC) \cite{Kull2017} using the calibration framework provided by \cite{Kueppers2020}. We replace the ML estimation by an SVI optimizer provided by Pyro \cite{Bingham2018} to infer calibration parameter distributions.
The \textit{histogram binning} (HB) method is also used as a reference model but without epistemic uncertainty modelling. All code is open source and available at \textit{https://github.com/EFS-OpenSource/calibration-framework}.

We use different subsets of the output data for calibration and for D-ECE calculation. This has the advantage of examining the performance of the calibration methods for different features (e.g. position and/or scale) or for different kinds of data distributions. Furthermore, for some applications (e.g. anchor-free models \cite{zhu2019}) only a subset of the actually used bounding box data is available or even relevant for the evaluation. Therefore, we either
use confidence $\pred$ only with $\numbins=20$ bins. Further, we add position $(\pred, \centerx, \centery)$ or shape $(\pred, \width, \height)$ information using $\numbins_\numcollectedvariates=8$ bins in each direction. Finally, we use all available information within calibration using $\numbins_\numcollectedvariates=5$ bins in each direction. To increase robustness of the D-ECE calculation, bins with less than 8 samples are neglected.

All experiments are restricted to the prediction and calibration of class \textit{pedestrians} only. We evaluate our methods on the MS COCO validation dataset \cite{Lin2014} consisting of 5,000 images with 36,781 annotated pedestrians.
We use a pretrained Faster R-CNN X101-FPN \cite{Ren2015} (14,487 predictions in total) and a pretrained RetinaNet R101-FPN \cite{Lin2017} (14,181 predictions) provided by Detectron2 \cite{Wu2019}. 
Furthermore, we also utilize the Cityscapes validation dataset \cite{Cordts2016} consisting of 500 images and 10,655 annotated pedestrians using the bounding box predictions of a pretrained Mask R-CNN R50-FPN \cite{He2017, Wu2019} with 3,462 predictions.

Similar to the training of a neural network, all calibration methods require a dedicated training set. 
Since no labels are available for the test data sets, our investigations are limited to the validation sets of each database. Thus, we splitted these sets randomly into training set (70\%) for building the calibration mapping and into test set (30\%) for evaluation. We repeated this 20 times to obtain an average result (with fixed seeds for reproducibility).\\

\textbf{Calibration evaluation.}
%Similar to \cite{Kueppers2020}, the D-ECE is calculated by means of different features given by the outputs of an object detector.
We compare the calibration results obtained by SVI with the standard models built by ML estimate. The results are given in Tab. \ref{result:table:calibration}.
We observe that the calibration performance of both methods is almost equal in nearly any case.
Using SVI, the mean values of the variational distributions converges to the ML estimates. 
Minor differences of the D-ECE may also result from inaccuracies of the binning scheme.
Similar to the experiments of \cite{Kueppers2020}, we can confirm that logistic and beta calibration have superior performance compared to histogram binning. This also holds for the SVI estimated models.
Therefore, we conclude that Bayesian confidence calibration offers the same performance compared to the standard ML estimation. However, using SVI has the advantage of also obtaining an uncertainty quantification for each result. Thus, we can provide an extended calibration approach without any loss of calibration performance.\\

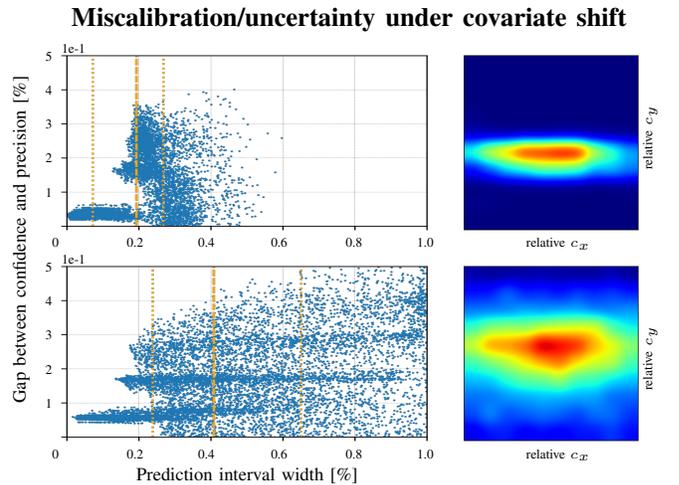
\begin{figure}[b!]
	\centering
	\begin{overpic}[tics=5, width=1.0\linewidth]{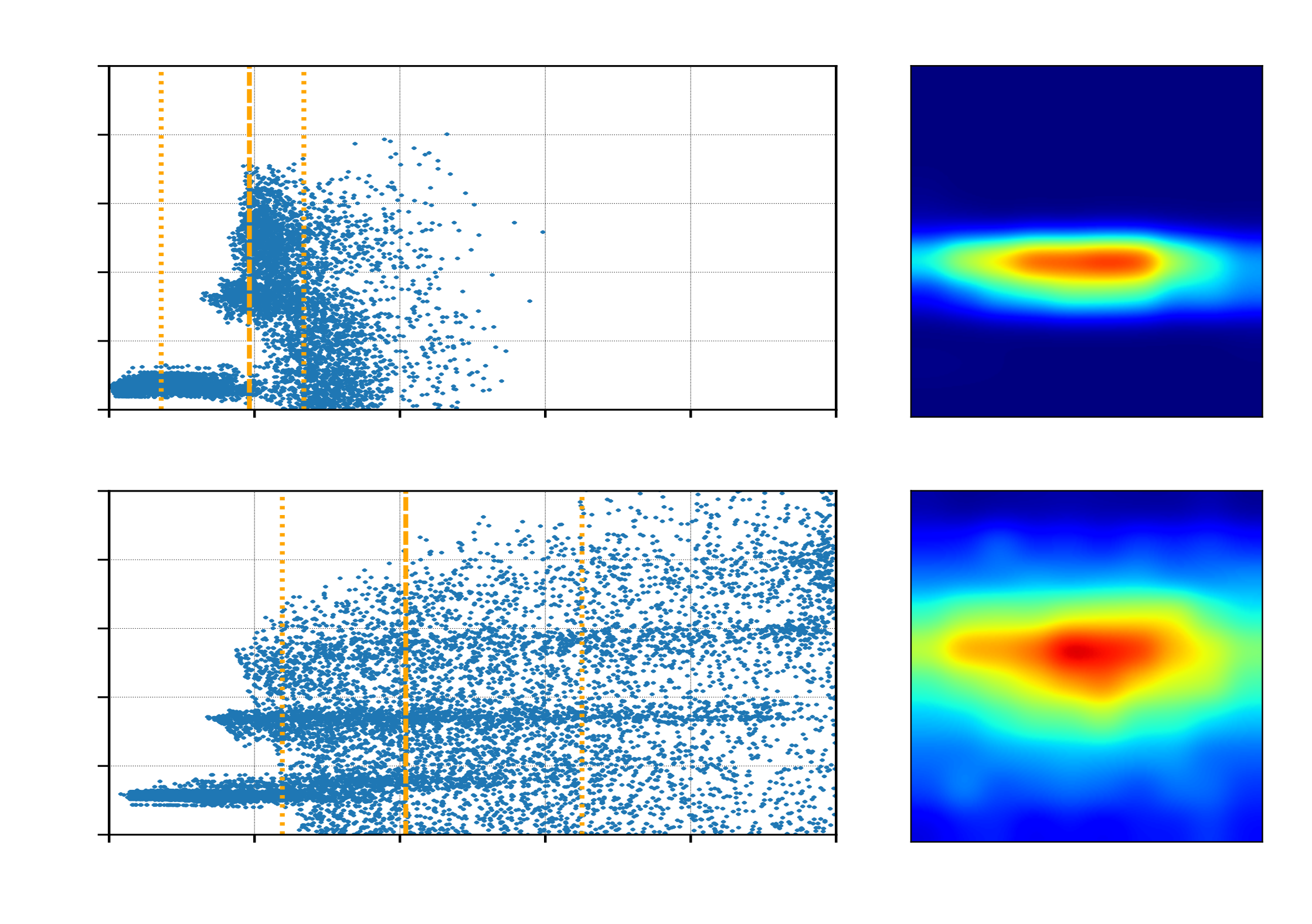}
		
		\put(9, 70){\textbf{Miscalibration/uncertainty under covariate shift}}
		\put(0, 12){\rotatebox{90}{\scriptsize{Gap between confidence and precision [\%]}}}
		\put(19, 0){\scriptsize{Prediction interval width [\%]}}
		
		\put(6,36){\tiny{0}}
		\put(17.5,36){\tiny{0.2}}
		\put(28.5,36){\tiny{0.4}}
		\put(39.7,36){\tiny{0.6}}
		\put(51.2,36){\tiny{0.8}}
		\put(62.2,36){\tiny{1.0}}
		\put(7.5, 66){\tiny{1e-1}}
		
		\put(6,3.5){\tiny{0}}
		\put(17.5,3.5){\tiny{0.2}}
		\put(28.5,3.5){\tiny{0.4}}
		\put(39.7,3.5){\tiny{0.6}}
		\put(51.2,3.5){\tiny{0.8}}
		\put(62.2,3.5){\tiny{1.0}}
		\put(7.5, 33.5){\tiny{1e-1}}
		
		\put(5,43.75){\tiny{1}}
		\put(5,49){\tiny{2}}
		\put(5,54.25){\tiny{3}}
		\put(5,59.5){\tiny{4}}
		\put(5,64.75){\tiny{5}}
		
		\put(5,11.25){\tiny{1}}
		\put(5,16.5){\tiny{2}}
		\put(5,21.75){\tiny{3}}
		\put(5,27){\tiny{4}}
		\put(5,32.25){\tiny{5}}
		
		\put(79,36){\tiny{relative $\centerx$}}
		\put(79,3.5){\tiny{relative $\centerx$}}
		\put(97,48){\rotatebox{90}{\tiny{relative $\centery$}}}
		\put(97,14){\rotatebox{90}{\tiny{relative $\centery$}}}
	\end{overpic}
	\caption{We use a Mask R-CNN trained on Cityscapes and build a calibration mapping (top row) with a certain data distribution (right column). Afterwards, we use the same network but on the MS COCO dataset (bottom row) with a different data distribution (right column) and apply the trained calibration mapping. Each point (left column) represents a prediction with an uncertainty interval and the respective calibration error. The orange dotted lines divide the data into the \{25, 50, 75\} percentiles. On the one hand, we observe an increasing calibration error as the prediction interval width increases. On the other hand, the prediction interval width can be used as a sufficient criterion if samples are out-of-distribution during inference.
	}
	\label{img:result:shift_distribution}
\end{figure}

\textbf{Uncertainty evaluation \& covariate shift.}
The epistemic uncertainty is evaluated by the PICP and MPIW scores. 
The results are shown in Tab. \ref{result:table:picp_mpiw}.
\begin{table*}[h]
	\centering
	\caption{Prediction interval coverage probability (PICP) for the 95\% prediction interval and mean prediction interval width (MPIW) for conditional dependent logistic calibration (LC) and beta calibration (BC) \cite{Kueppers2020} using SVI estimation. The structure of each subtable is equal to Tab. \ref{result:table:calibration}.}
	\label{result:table:picp_mpiw}
	\caption*{PICP [\%] on MS COCO validation set \cite{Lin2014} with IoU 0.50 (left) and 0.75 (right)}
	\begin{subtable}{0.48\linewidth}
		\centering
		\caption{Faster R-CNN X101-FPN \cite{Ren2015} with IoU 0.50}
		\resizebox{0.95\linewidth}{!}{%
			\begin{tabular}{llrrrr}
\toprule
     &    &  $(\pred)$ &  $(\pred, \centerx, \centery)$ &    $(\pred, \width, \height)$ &    full \\
\midrule
\multirow{2}{*}{PICP} & LC &  84.719 &  78.326 &  83.471 &  82.217 \\
     & BC &  79.431 &  80.051 &  84.315 &  80.627 \\
\cline{1-6}
\multirow{2}{*}{MPIW} & LC &   5.763 &  14.793 &  13.458 &  21.248 \\
     & BC &   7.026 &  15.340 &  16.527 &  20.193 \\
\bottomrule
\end{tabular}

		}
	\end{subtable}
	\vspace{1em}
	\begin{subtable}{0.48\linewidth}
		\centering
		\caption{Faster R-CNN X101-FPN \cite{Ren2015} with IoU 0.75}
		\resizebox{0.95\linewidth}{!}{%
			\begin{tabular}{llrrrr}
\toprule
     &    &  $(\pred)$ &  $(\pred, \centerx, \centery)$ &    $(\pred, \width, \height)$ &    full \\
\midrule
\multirow{2}{*}{PICP} & LC &  99.161 &  95.455 &  96.292 &  96.250 \\
     & BC &  85.824 &  97.340 &  97.272 &  95.510 \\
\cline{1-6}
\multirow{2}{*}{MPIW} & LC &   6.947 &  17.655 &  16.551 &  25.724 \\
     & BC &   6.979 &  19.120 &  20.172 &  24.721 \\
\bottomrule
\end{tabular}

		}
	\end{subtable}
	
	\begin{subtable}{0.48\linewidth}
		\centering
		\caption{RetinaNet R101-FPN \cite{Lin2017} with IoU 0.50}
		\resizebox{0.95\linewidth}{!}{%
			\begin{tabular}{llrrrr}
\toprule
     &    &  $(\pred)$ &  $(\pred, \centerx, \centery)$ &    $(\pred, \width, \height)$ &    full \\
\midrule
\multirow{2}{*}{PICP} & LC &  82.237 &  78.886 &  86.076 &  82.712 \\
     & BC &  80.022 &  77.568 &  75.667 &  72.555 \\
\cline{1-6}
\multirow{2}{*}{MPIW} & LC &   6.324 &  15.187 &  15.501 &  22.334 \\
     & BC &   8.148 &  17.048 &  15.641 &  19.362 \\
\bottomrule
\end{tabular}

		}
	\end{subtable}
	\vspace{1.5em}
	\begin{subtable}{0.48\linewidth}
		\centering
		\caption{RetinaNet R101-FPN \cite{Lin2017} with IoU 0.75}
		\resizebox{0.95\linewidth}{!}{%
			\begin{tabular}{llrrrr}
\toprule
     &    &  $(\pred)$ &  $(\pred, \centerx, \centery)$ &    $(\pred, \width, \height)$ &    full \\
\midrule
\multirow{2}{*}{PICP} & LC &  98.764 &  95.448 &  98.465 &  97.195 \\
     & BC &  97.404 &  96.206 &  92.324 &  89.598 \\
\cline{1-6}
\multirow{2}{*}{MPIW} & LC &   7.021 &  18.644 &  17.963 &  26.962 \\
     & BC &   8.954 &  20.124 &  19.458 &  24.093 \\
\bottomrule
\end{tabular}

		}
	\end{subtable}
	
	\caption*{PICP [\%] on Cityscapes validation set \cite{Cordts2016} with IoU 0.50 (left) and 0.75 (right)}
	\begin{subtable}{0.48\linewidth}
		\centering
		\caption{Mask-RCNN R50-FPN \cite{He2017} with IoU 0.50}
		\resizebox{0.95\linewidth}{!}{%
			\begin{tabular}{llrrrr}
\toprule
     &    &  $(\pred)$ &  $(\pred, \centerx, \centery)$ &    $(\pred, \width, \height)$ &    full \\
\midrule
\multirow{2}{*}{PICP} & LC &  89.062 &  89.560 &  92.652 &  84.600 \\
     & BC &  74.678 &  91.162 &  93.628 &  85.545 \\
\cline{1-6}
\multirow{2}{*}{MPIW} & LC &   6.461 &  16.969 &  16.762 &  22.783 \\
     & BC &   7.083 &  16.207 &  18.869 &  22.217 \\
\bottomrule
\end{tabular}

		}
	\end{subtable}
	\begin{subtable}{0.48\linewidth}
		\centering
		\caption{Mask-RCNN R50-FPN \cite{He2017} with IoU 0.75}
		\resizebox{0.95\linewidth}{!}{%
			\begin{tabular}{llrrrr}
\toprule
     &    &  $(\pred)$ &  $(\pred, \centerx, \centery)$ &    $(\pred, \width, \height)$ &    full \\
\midrule
\multirow{2}{*}{PICP} & LC &  98.848 &  99.135 &  99.217 &  98.576 \\
     & BC &  95.166 &  99.110 &  99.315 &  98.321 \\
\cline{1-6}
\multirow{2}{*}{MPIW} & LC &   7.664 &  20.202 &  19.624 &  28.443 \\
     & BC &   6.901 &  19.660 &  21.642 &  25.141 \\
\bottomrule
\end{tabular}

		}
	\end{subtable}
\end{table*}
We achieve reasonable uncertainty estimates for the prediction interval using SVI with PICP scores close to $95\%$ in many cases.
Further, increasing MPIW scores are observed as the number of dimensions used for calibration increases.
We assume that the increasing MPIW score is a result of the larger data space while using the same amount of samples.
In contrast to aleatoric uncertainty, the epistemic uncertainty can be minimized given more data \cite{Kendall2017}. Therefore, it would be interesting to investigate if the uncertainty decreases when more data is used. So far this is not feasible with any publicly available data base for object detection.

%Interestingly, the interval estimates gets better as the IoU score increases. We suspect that the predicted uncertainty is related to precision, which decreases for a higher IoU score. 
Furthermore, we observe that epistemic uncertainty is highly correlated with the data distribution used for calibration training. This is demonstrated in Fig. \ref{img:result:shift_distribution} where we measure the uncertainty of a calibration mapping for MS COCO predictions by a Mask R-CNN, that has been trained and also calibrated on Cityscapes. 
Since COCO images exhibit considerably more diversity than Cityscapes, we have several predictions that have not been covered by the sample distribution during calibration training. Samples in sparsely populated regions thus have a significantly higher prediction interval width. Therefore, we can use the prediction interval width as a sufficient criterion to detect samples that are out-of-calibration-training distribution during inference.

In conclusion, we demonstrate that it is possible to achieve state-of-the-art calibration performance using Bayesian confidence calibration. We further observe only minor differences between logistic and beta calibration. Most important, we provide a framework to obtain calibrated confidence estimates in conjunction with qualitatively good estimates for epistemic uncertainty. This uncertainty might additionally be used as a sufficient criterion to detect a possible covariate shift. Therefore, this framework is particularly suitable for safety-critical applications where a reliable uncertainty quantification is of special interest.

\section{Conclusion}
In this paper we present a novel Bayesian framework for confidence calibration to quantify epistemic uncertainty within a calibration mapping. We extend common calibration methods for object detection \cite{Kueppers2020} by stochastic variational inference (SVI) to infer distributions as the calibration parameters. Therefore, it is possible to obtain a sample distribution as the calibrated confidence estimate. This allows for a quantification of the calibration mapping's intrinsic uncertainty.
In our experiments we show that our framework achieves state-of-the-art calibration performance on the detection expected calibration error (D-ECE) compared to the commonly used ML-estimated models. We further evaluate the epistemic uncertainty and show that our framework provides meaningful prediction intervals that cover the observed frequency in most cases on the one hand. On the other hand, the uncertainty turned out to be a sufficient indicator of a possible covariate shift between calibration training data and testing set.

In addition to the confidence estimation, it is possible to also use the epistemic uncertainty even to non-BNNs for calibration verification. Our framework is therefore useful especially for safety-critical applications such as driver assistance systems or medical diagnosis where a reliable confidence indication is of major significance.

\section*{Acknowledgement}
The authors gratefully acknowledge support of this work by Elektronische Fahrwerksysteme GmbH, Gaimersheim, Germany. The research leading to the results presented above are funded by the German Federal Ministry for Economic Affairs and Energy within the project “KI Absicherung – Safe AI for automated driving".

%%%%%%%%%%%%%%%%%%%%%%%%%%%%%%%%%%%%%%%%%%%%%%%%%%%%%%%%%%%%%%%%%%%%%%%%%%%%%%%%

%%%%%%%%%%%%%%%%%%%%%%%%%%%%%%%%%%%%%%%%%%%%%%%%%%%%%%%%%%%%%%%%%%%%%%%%%%%%%%%%

%%%%%%%%%%%%%%%%%%%%%%%%%%%%%%%%%%%%%%%%%%%%%%%%%%%%%%%%%%%%%%%%%%%%%%%%%%%%%%%%

%%%%%%%%%%%%%%%%%%%%%%%%%%%%%%%%%%%%%%%%%%%%%%%%%%%%%%%%%%%%%%%%%%%%%%%%%%%%%%%%

{\small
	\bibliographystyle{IEEEtran}
	\bibliography{bib/bibliography}

\begin{thebibliography}{10}
\providecommand{\url}[1]{#1}
\csname url@rmstyle\endcsname
\providecommand{\newblock}{\relax}
\providecommand{\bibinfo}[2]{#2}
\providecommand\BIBentrySTDinterwordspacing{\spaceskip=0pt\relax}
\providecommand\BIBentryALTinterwordstretchfactor{4}
\providecommand\BIBentryALTinterwordspacing{\spaceskip=\fontdimen2\font plus
\BIBentryALTinterwordstretchfactor\fontdimen3\font minus
  \fontdimen4\font\relax}
\providecommand\BIBforeignlanguage[2]{{%
\expandafter\ifx\csname l@#1\endcsname\relax
\typeout{** WARNING: IEEEtran.bst: No hyphenation pattern has been}%
\typeout{** loaded for the language `#1'. Using the pattern for}%
\typeout{** the default language instead.}%
\else
\language=\csname l@#1\endcsname
\fi
#2}}

\bibitem{Guo2018}
C.~Guo, G.~Pleiss, Y.~Sun, and K.~Q. Weinberger, ``{On Calibration of Modern
  Neural Networks},'' in \emph{Proceedings of the 34th International Conference
  on Machine Learning}, ser. Proceedings of Machine Learning Research,
  vol.~70.\hskip 1em plus 0.5em minus 0.4em\relax PMLR, August 2017, pp.
  1321--1330.

\bibitem{Niculescu2005}
A.~Niculescu-Mizil and R.~Caruana, ``Predicting good probabilities with
  supervised learning,'' in \emph{Proceedings of the 22nd International
  Conference on Machine Learning}, 2005, pp. 625--632.

\bibitem{Naeini2015}
M.~Naeini, G.~Cooper, and M.~Hauskrecht, ``{Obtaining Well Calibrated
  Probabilities Using Bayesian Binning},'' in \emph{Proceedings of the 29th
  AAAI Conference on Artificial Intelligence}, 2015, pp. 2901--2907.

\bibitem{Kueppers2020}
F.~Küppers, J.~Kronenberger, A.~Shantia, and A.~Haselhoff, ``Multivariate
  confidence calibration for object detection,'' in \emph{Proceedings of the
  IEEE/CVF Conference on Computer Vision and Pattern Recognition Workshops},
  2020, pp. 326--327.

\bibitem{Kull2017}
M.~Kull, T.~Silva~Filho, and P.~Flach, ``Beta calibration: a well-founded and
  easily implemented improvement on logistic calibration for binary
  classifiers,'' in \emph{Artificial Intelligence and Statistics}, 2017, pp.
  623--631.

\bibitem{Neumann2018}
L.~Neumann, A.~Zisserman, and A.~Vedaldi, ``{Relaxed Softmax: Efficient
  Confidence Auto-Calibration for Safe Pedestrian Detection},'' in
  \emph{Workshop on Machine Learning for Intelligent Transportation Systems
  (NIPS)}, 2018.

\bibitem{Kuleshov2018}
V.~Kuleshov, N.~Fenner, and S.~Ermon, ``{Accurate Uncertainties for Deep
  Learning Using Calibrated Regression},'' in \emph{International Conference on
  Machine Learning (ICML)}, 2018, pp. 2801--2809.

\bibitem{Song2018}
H.~Song, M.~Kull, and P.~Flach, ``Non-parametric calibration of probabilistic
  regression,'' \emph{CoRR}, 2018.

\bibitem{Pearce2018}
T.~Pearce, A.~Brintrup, M.~Zaki, and A.~Neely, ``High-quality prediction
  intervals for deep learning: A distribution-free, ensembled approach,'' in
  \emph{International Conference on Machine Learning}, 2018, pp. 4075--4084.

\bibitem{Song2019}
H.~Song, T.~Diethe, M.~Kull, and P.~Flach, ``Distribution calibration for
  regression,'' in \emph{Proceedings of the 36th International Conference on
  Machine Learning}, ser. Proceedings of Machine Learning Research,
  K.~Chaudhuri and R.~Salakhutdinov, Eds., vol.~97.\hskip 1em plus 0.5em minus
  0.4em\relax Long Beach, California, USA: PMLR, 09--15 Jun 2019, pp.
  5897--5906.

\bibitem{Ding2020}
Y.~Ding, J.~Liu, J.~Xiong, and Y.~Shi, ``Revisiting the evaluation of
  uncertainty estimation and its application to explore model
  complexity-uncertainty trade-off,'' in \emph{Proceedings of the IEEE/CVF
  Conference on Computer Vision and Pattern Recognition Workshops}, 2020, pp.
  4--5.

\bibitem{Platt1999}
J.~Platt, ``{Probabilistic Outputs for Support Vector Machines and Comparisons
  to Regularized Likelihood Methods},'' \emph{Advances in Large Margin
  Classifiers}, pp. 61--74, 1999.

\bibitem{Zadrozny2001}
B.~Zadrozny and C.~Elkan, ``{Obtaining Calibrated Probability Estimates from
  Decision Trees and Naive Bayesian Classifiers},'' in \emph{Proceedings of the
  Eighteenth International Conference on Machine Learning (ICML)}, 2001, pp.
  609--616.

\bibitem{Feng2019}
D.~Feng, L.~Rosenbaum, C.~Glaeser, F.~Timm, and K.~Dietmayer, ``Can we trust
  you? on calibration of a probabilistic object detector for autonomous
  driving,'' \emph{arXiv preprint}, 2019.

\bibitem{Mukhoti2020}
J.~Mukhoti, V.~Kulharia, A.~Sanyal, S.~Golodetz, P.~H. Torr, and P.~K. Dokania,
  ``Calibrating deep neural networks using focal loss,'' in \emph{Advances in
  Neural Information Processing Systems}, 2020.

\bibitem{Pereyra2017}
G.~Pereyra, G.~Tucker, J.~Chorowski, Łukasz Kaiser, and G.~Hinton,
  ``Regularizing neural networks by penalizing confident output
  distributions,'' \emph{CoRR}, 2017.

\bibitem{Seo2019}
S.~Seo, P.~H. Seo, and B.~Han, ``{Learning for Single-Shot Confidence
  Calibration in Deep Neural Networks Through Stochastic Inferences},'' in
  \emph{The IEEE Conference on Computer Vision and Pattern Recognition (CVPR)},
  June 2019.

\bibitem{Graves2011}
\BIBentryALTinterwordspacing
A.~Graves, ``Practical variational inference for neural networks,'' in
  \emph{Advances in Neural Information Processing Systems 24}, J.~Shawe-Taylor,
  R.~S. Zemel, P.~L. Bartlett, F.~Pereira, and K.~Q. Weinberger, Eds.\hskip 1em
  plus 0.5em minus 0.4em\relax Curran Associates, Inc., 2011, pp. 2348--2356.
  [Online]. Available:
  \url{http://papers.nips.cc/paper/4329-practical-variational-inference-for-neural-networks.pdf}
\BIBentrySTDinterwordspacing

\bibitem{Blundell2015}
C.~Blundell, J.~Cornebise, K.~Kavukcuoglu, and D.~Wierstra, ``Weight
  uncertainty in neural networks,'' in \emph{Proceedings of the 32nd
  International Conference on International Conference on Machine Learning -
  Volume 37}, ser. ICML’15.\hskip 1em plus 0.5em minus 0.4em\relax JMLR.org,
  2015, p. 1613–1622.

\bibitem{Gal2016}
Y.~Gal and Z.~Ghahramani, ``{Dropout as a Bayesian Approximation: Representing
  Model Uncertainty in Deep Learning},'' in \emph{International Conference on
  Machine Learning (ICML)}, 2016, pp. 1050--1059.

\bibitem{Kendall2017}
A.~Kendall and Y.~Gal, ``{What Uncertainties Do We Need in Bayesian Deep
  Learning for Computer Vision?}'' in \emph{Advances in Neural Information
  Processing Systems (NIPS)}, 2017, pp. 5574--5584.

\bibitem{Lakshminarayanan2017}
B.~Lakshminarayanan, A.~Pritzel, and C.~Blundell, ``Simple and scalable
  predictive uncertainty estimation using deep ensembles,'' in \emph{Advances
  in neural information processing systems}, 2017, pp. 6402--6413.

\bibitem{Hall2020}
D.~Hall, F.~Dayoub, J.~Skinner, H.~Zhang, D.~Miller, P.~Corke, G.~Carneiro,
  A.~Angelova, and N.~S{\"u}nderhauf, ``Probabilistic object detection:
  Definition and evaluation,'' in \emph{The IEEE Winter Conference on
  Applications of Computer Vision}, 2020, pp. 1031--1040.

\bibitem{Jiang2018}
B.~Jiang, R.~Luo, J.~Mao, T.~Xiao, and Y.~Jiang, ``Acquisition of localization
  confidence for accurate object detection,'' in \emph{Proceedings of the
  European Conference on Computer Vision (ECCV)}, 2018, pp. 784--799.

\bibitem{Rezatofighi2019}
H.~Rezatofighi, N.~Tsoi, J.~Gwak, A.~Sadeghian, I.~Reid, and S.~Savarese,
  ``Generalized intersection over union: A metric and a loss for bounding box
  regression,'' in \emph{Proceedings of the IEEE Conference on Computer Vision
  and Pattern Recognition}, 2019, pp. 658--666.

\bibitem{Allikivi2019}
M.-L. Allikivi and M.~Kull, ``Non-parametric bayesian isotonic calibration:
  Fighting over-confidence in binary classification,'' in \emph{Joint European
  Conference on Machine Learning and Knowledge Discovery in Databases}.\hskip
  1em plus 0.5em minus 0.4em\relax Springer, 2019, pp. 103--120.

\bibitem{Jordan1999}
M.~I. Jordan, Z.~Ghahramani, T.~S. Jaakkola, and L.~K. Saul, ``An introduction
  to variational methods for graphical models,'' \emph{Machine learning},
  vol.~37, no.~2, pp. 183--233, 1999.

\bibitem{Hoffman2013}
\BIBentryALTinterwordspacing
M.~D. Hoffman, D.~M. Blei, C.~Wang, and J.~Paisley, ``Stochastic variational
  inference,'' \emph{Journal of Machine Learning Research}, vol.~14, no.~4, pp.
  1303--1347, 2013. [Online]. Available:
  \url{http://jmlr.org/papers/v14/hoffman13a.html}
\BIBentrySTDinterwordspacing

\bibitem{Gal2016a}
Y.~Gal, ``Uncertainty in deep learning,'' \emph{University of Cambridge},
  vol.~1, p.~3, 2016.

\bibitem{Rueda2007}
M.~Rueda, S.~Mart{\'\i}nez-Puertas, H.~Mart{\'\i}nez-Puertas, and A.~Arcos,
  ``Calibration methods for estimating quantiles,'' \emph{Metrika}, vol.~66,
  no.~3, pp. 355--371, 2007.

\bibitem{Gneiting2007}
T.~Gneiting, F.~Balabdaoui, and A.~E. Raftery, ``Probabilistic forecasts,
  calibration and sharpness,'' \emph{Journal of the Royal Statistical Society:
  Series B (Statistical Methodology)}, vol.~69, no.~2, pp. 243--268, 2007.

\bibitem{Fasiolo2020}
\BIBentryALTinterwordspacing
M.~Fasiolo, S.~N. Wood, M.~Zaffran, R.~Nedellec, and Y.~Goude, ``Fast
  calibrated additive quantile regression,'' \emph{Journal of the American
  Statistical Association}, vol.~0, no.~0, pp. 1--11, 2020. [Online].
  Available: \url{https://doi.org/10.1080/01621459.2020.1725521}
\BIBentrySTDinterwordspacing

\bibitem{Lin2014}
T.-Y. Lin, M.~Maire, S.~Belongie, J.~Hays, P.~Perona, D.~Ramanan,
  P.~Doll{\'a}r, and C.~L. Zitnick, ``{Microsoft COCO: Common objects in
  context},'' in \emph{European Conference on Computer Vision (ECCV)}.\hskip
  1em plus 0.5em minus 0.4em\relax Springer, 2014, pp. 740--755.

\bibitem{Ren2015}
S.~Ren, K.~He, R.~Girshick, and J.~Sun, ``{Faster R-CNN: Towards real-time
  object detection with region proposal networks},'' in \emph{Advances in
  Neural Information Processing Systems (NIPS)}, 2015, pp. 91--99.

\bibitem{Lin2017}
T.-Y. Lin, P.~Goyal, R.~Girshick, K.~He, and P.~Doll{\'a}r, ``{Focal Loss for
  Dense Object Detection},'' in \emph{Proceedings of the IEEE International
  Conference on Computer Vision (ICCV)}, 2017, pp. 2980--2988.

\bibitem{Cordts2016}
M.~Cordts, M.~Omran, S.~Ramos, T.~Rehfeld, M.~Enzweiler, R.~Benenson,
  U.~Franke, S.~Roth, and B.~Schiele, ``The cityscapes dataset for semantic
  urban scene understanding,'' in \emph{Proc. of the IEEE Conference on
  Computer Vision and Pattern Recognition (CVPR)}, 2016.

\bibitem{He2017}
K.~He, G.~Gkioxari, P.~Doll{\'a}r, and R.~Girshick, ``Mask r-cnn,'' in
  \emph{Proceedings of the IEEE international conference on computer vision},
  2017, pp. 2961--2969.

\bibitem{Bingham2018}
E.~Bingham, J.~P. Chen, M.~Jankowiak, F.~Obermeyer, N.~Pradhan, T.~Karaletsos,
  R.~Singh, P.~Szerlip, P.~Horsfall, and N.~D. Goodman, ``{Pyro: Deep Universal
  Probabilistic Programming},'' \emph{Journal of Machine Learning Research},
  2018.

\bibitem{zhu2019}
C.~Zhu, Y.~He, and M.~Savvides, ``Feature selective anchor-free module for
  single-shot object detection,'' in \emph{Proceedings of the IEEE Conference
  on Computer Vision and Pattern Recognition}, 2019, pp. 840--849.

\bibitem{Wu2019}
Y.~Wu, A.~Kirillov, F.~Massa, W.-Y. Lo, and R.~Girshick, ``Detectron2,''
  \url{https://github.com/facebookresearch/detectron2}, 2019.

\end{thebibliography}
}

\end{document}